\pdfoutput=1

\documentclass[11pt]{article}

\usepackage{emnlp2021}

\usepackage{times}
\usepackage{latexsym}
\usepackage{xspace}

\usepackage[T1]{fontenc}

\usepackage[utf8]{inputenc}

\usepackage{microtype}

\usepackage{stfloats}
\usepackage{amsmath}
\usepackage{multirow}
\usepackage{tikz}
\usetikzlibrary{topaths}
\tikzstyle{every picture}+=[remember picture,inner xsep=0,inner ysep=0.25ex]

\usepackage{textcomp}

\newcommand{\nsm}{A\textsc{na}.\xspace}
\newcommand{\am}{A\textsc{ll}\xspace}

\newif\ifcomments
\commentstrue
\ifcomments
    \providecommand{\matt}[1]{{\protect\color{teal}{\bf [Matt: #1]}}}
    \providecommand{\zhaofeng}[1]{{\protect\color{olive}{\bf [Zhaofeng: #1]}}}
    \providecommand{\markn}[1]{{\protect\color{blue}{\bf [Mark: #1]}}}
\else
    \providecommand{\matt}[1]{}
    \providecommand{\zhaofeng}[1]{}
    \providecommand{\markn}[1]{}
\fi

\title{Understanding Mention Detector-Linker Interaction\\ in Neural Coreference Resolution}

\author{Zhaofeng Wu \\
  Paul G. Allen School of CSE\\ University of Washington \\
  \texttt{zfw7@cs.washington.edu} \\\And
  Matt Gardner \\
  Allen Institute for AI \\
  \texttt{mattg@allenai.org} \\}

\date{}

\begin{document}
\maketitle
\begin{abstract}

Despite significant recent progress in coreference resolution, the quality of current state-of-the-art systems still considerably trails behind human-level performance. Using the CoNLL-2012 and PreCo datasets, we dissect the best instantiation of the mainstream end-to-end coreference resolution model that underlies most current best-performing coreference systems, and empirically analyze the behavior of its two components: mention detector and mention linker. While the detector traditionally focuses heavily on recall as a design decision, we demonstrate the importance of precision, calling for their balance. However, we point out the difficulty in building a precise detector due to its inability to make important anaphoricity decisions. We also highlight the enormous room for improving the linker and show that the rest of its errors mainly involve pronoun resolution. We propose promising next steps and hope our findings will help future research in coreference resolution.

\end{abstract}

\section{Introduction}

Coreference resolution identifies mentions in a document that co-refer to the same entity. It is an important task facilitating many applications such as reading comprehension~\cite{dasigi-etal-2019-quoref} and text summarization~\cite{azzam-etal-1999-using}.

\citet{lee-etal-2017-end} proposed the first neural end-to-end architecture for coreference resolution. Most recent systems use it as a backbone and employ better scoring functions~\cite{zhang-etal-2018-neural-coreference}, pruning procedures~\cite{lee-etal-2018-higher}, or token representations~\cite{joshi-etal-2019-bert, joshi2020spanbert}.\footnote{Except \citet{wu-etal-2020-corefqa} which has not seen wide adoption.} Despite this usage, little in-depth analysis has been done to better understand the inner workings of such an influential system. \citet{xu-choi-2020-revealing} analyzed the effect of the high-order inference, while \citet{subramanian-roth-2019-improving} and \citet{zhao-etal-2018-gender} respectively examined its generalizability and gender bias. Few work has inspected the interaction between its components. \citet{lu-ng-2020-conundrums} conducted oracle experiments that are related to ours, but without fine-grained control over confounding factors affecting oracle mentions. Such an understanding is important: for example, \citet{kummerfeld-klein-2013-error}'s dissection of the then-best classical coreference systems inspired many important follow-up works \cite[\emph{inter alia}]{peng-etal-2015-solving, martschat-strube-2015-latent, wiseman-etal-2016-learning}. However, it is unknown if observations on such classical feature-based and often pipelined systems extend to current neural end-to-end models.

We consider the best instantiation of this model family, SpanBERT~\cite{joshi2020spanbert} + c2f-coref~\cite{lee-etal-2018-higher}, and investigate the interaction between its two components: mention detector and mention linker. We study how their errors independently or jointly affect the final clustering.

Using the CoNLL-2012~\cite{pradhan2012conll} and PreCo~\cite{chen-etal-2018-preco} datasets, we highlight the low-precision, high-recall nature of the detector. While traditionally only recall is emphasized for the detector as a design decision (\citealp{lee2011stanford}; \citealp{lee-etal-2017-end}), we show huge degradation from noisy mentions and, perhaps surprisingly, increasing the number of candidates considered by the baseline linker only deteriorates the performance. While some classical coreference pipelines focused on detector precision~\cite{Uryupina2009DetectingAA}, it is rarely emphasized for current end-to-end systems. We hence stress the importance of a precision-recall balance for the detector and demonstrate how pruning hyperparameters, in addition to reducing computational complexity, control this trade-off. However, we show the difficulty of obtaining a precise detector by demonstrating the importance of anaphoricity decisions and the inability of the detector to make them. Finally, we highlight the high potential of the linker and that the remaining errors mainly involve pronoun resolution. We hope this work sheds light on the internals of the mainstream coreference system and, with our proposed next steps, catalyze future research. We believe some of our findings may also transfer to other tasks with a similar joint span detection and span (pair) classification architecture, such as SRL~\cite{he-etal-2018-jointly}, IE~\cite{luan-etal-2019-general}, and entity linking~\cite{kolitsas-etal-2018-end}.  See \citet{jiang-etal-2020-generalizing} which subsumes many other tasks under such a span-based framework.

\section{Background}

\paragraph{Model} We study the coarse-to-fine coreference system (c2f-coref; \citealt{lee-etal-2018-higher}). It assigns an antecedent for every span in a document of length $T$, including a dummy that indicates non-mentions or non-anaphoric mentions. The final clustering is the transitive closure of connected spans.
The system consists of a mention detector and a mention linker. The detector scores all $O(T^2)$ spans up to length $L$ and outputs the $\lambda T$ highest-scoring spans as possibly anaphoric mentions. The linker links each mention candidate with the highest-scoring antecedent among $K$ ones. Hyperparameters $L$, $\lambda$, and $K$ control the number of considered spans and antecedents, reducing computational complexity.

\paragraph{Data} CoNLL-2012 is the most common dataset to test coreference models. However, it lacks singleton mention annotation~\cite{pradhan2012conll}.

Singleton, or non-anaphoric, mentions do not co-refer with other spans, e.g. ``The dog'' in ``[The dog] barks.'' However, they may become anaphoric in another context, e.g. ``[The dog] barks at [itself].'' Being a mention is a span's inherent property, while anaphoricity, whether or not a mention co-refers, is context-dependent. We use ``all mentions'' to refer to the union of singleton and anaphoric mentions.

To understand the effect of singleton mentions, we heuristically generate all mentions for CoNLL-12 (\S\ref{sec:heuristics}) for relevant experiments. We also experiment with PreCo, a coreference dataset with annotated singleton mentions.  We do all analyses on development sets and report dataset statistics in \S\ref{sec:dataset-statistics}.

\section{Experiments}

\paragraph{Settings} We embed tokens with SpanBERT-large, a pre-trained transformer~\cite{vaswani2017attention} with state-of-the-art performance in coreference resolution.
We choose $L = 30,\lambda = 0.4,K = 50$.
We only keep the first 110 sentences per document during training. To reduce confounding factors, we do not use speaker and genre metadata.%

\paragraph{``Original'' System} refers to a standard SpanBERT + c2f-coref trained baseline. Its F$_1$ score\footnote{We use coreference F$_1$ to refer to the average F$_1$ of MUC, B$^3$, and CEAF$_{\phi_4}$, the most common coreference metric.} is reported in Table~\ref{tab:original}, similar to the results in \citet{joshi2020spanbert} considering we disregard metadata.

\begin{table}[t]
\centering
\begin{tabular}{c|cc}

\hline
& CoNLL-12 & PreCo \\ \hline
Coref F$_1$ & 79.17 & 85.04 \\ \hline
\nsm P & 28.37 & 39.23 \\
\nsm R & 96.42 & 98.40 \\
\am P & 82.04 & 76.55 \\
\am R & 57.35 & 95.98 \\ \hline

\end{tabular}
\caption{\label{tab:original} Original system coreference F$_1$ and precision / recall for anaphoric mentions (\nsm) and all mentions (\am) on CoNLL-12 and PreCo development sets.}
\vspace{-0.3cm}
\end{table}

\paragraph{Oracles} We build oracle detectors where, starting from the original system's mention candidates (its detector output), we either remove all non-gold mentions (prefect precision), add all missing gold mentions (perfect recall), or both (perfect precision \& recall). We give the altered, rather than the original, mention candidates to the linker. We consider both anaphoric mentions and all mentions as gold mentions and modify either in a post-hoc manner or re-train the system with the altered candidates.
To control for a non-trainable detector, we train only a linker reusing the original system's mention candidates, dubbed \textbf{Fixed Detector}. We consider this baseline as the comparison target for the oracles. Besides oracle detectors, we also build an oracle linker that assigns the correct antecedent (including dummy) to each of the $\lambda T$ mention candidates.

\section{Precision-Recall Trade-Off for the Mention Detector}
\label{sec:pr-re-analysis}

\begin{table*}[t]
\centering
\begin{tabular}{c|l|c|cccccc}
\hline
\multicolumn{2}{c|}{} & Span & Conflated & Extra & Extra & Divided & Missing & Missing \\
\multicolumn{2}{c|}{} & Error & Entities & Mention & Entity & Entity & Mention & Entity \\ \hline
\multicolumn{2}{c|}{Original} & 1.6 & 3.1 & 1.3 & 3.1 & 2.6 & 2.1 & 4.5 \\ \hline
\multicolumn{2}{c|}{Fixed Detector} & 1.7 & 3.1 & 1.7 & 3.3 & 2.8 & 2.1 & 4.7 \\ \hline

\nsm & Perfect P & 0.0 & 2.4 & 0.0 & 0.0 & 2.0 & 2.6 & 5.7 \\
Post-hoc & Perfect R & 1.6 & 3.1 & 1.4 & 3.1 & 2.6 & 2.0 & 4.4 \\
Oracle & Perfect P\&R & 0.0 & 2.3 & 0.0 & 0.0 & 1.9 & 2.5 & 5.5 \\ \hline

\nsm & Perfect P & 0.0 & 3.4 & 0.0 & 0.0 & 0.9 & 1.5 & 1.4 \\
Re-train & Perfect R & 0.7 & 3.1 & 2.0 & 3.4 & 2.6 & 1.5 & 4.4 \\
Oracle & Perfect P\&R & 0.0 & 3.0 & 0.0 & 0.0 & 1.0 & 0.4 & 0.3 \\ \hline

\end{tabular}
\caption{\label{tab:fix-errors} The F$_1$ score improvement after fixing different types of errors on the CoNLL-12 development set. The errors are independently fixed after span errors are fixed. The categorization is from \citet{kummerfeld-klein-2013-error}.}
\vspace{-0.3cm}
\end{table*}

\begin{table}[t]
\centering
\begin{tabular}{c|l|cc}
\hline
\multicolumn{2}{c|}{} & CoNLL-12 & PreCo \\ \hline

\multicolumn{2}{c|}{Fixed Detector} & 78.28 & 84.64 \\ \hline

\nsm & Perfect P & 86.02 & 90.31 \\
Post-hoc & Perfect R & 79.37 & 85.17 \\
Oracle & Perfect P\&R & 86.28 & 90.45 \\ \hline

\nsm & Perfect P & 89.98 & 95.09 \\
Re-train & Perfect R & 79.65 & 85.22 \\
Oracle & Perfect P\&R & 92.39 & 96.50 \\ \hline

\am & Perfect P & 79.48 & 88.37 \\
Re-train & Perfect R & 78.52 & 85.23 \\
Oracle & Perfect P\&R & 80.05 & 89.13 \\ \hline

\multicolumn{2}{c|}{Oracle Linker} & 97.07 & 98.69 \\ \hline

\end{tabular}
\caption{\label{tab:oracles} Baseline and oracle coreference F$_1$ for anaphoric mentions (\nsm) and all mentions (\am) on CoNLL-12 and PreCo development sets. ``Fixed Detector'' is the baseline with a non-trainable detector. The middle three sections are oracle detectors with perfect candidate precision/recall. The last row is an oracle linker that always makes correct antecedent decisions.}
\vspace{-0.3cm}
\end{table}

Traditionally, coreference systems heavily favor recall over precision for the detector ~\cite{lee2011stanford} as the linker cannot recover missed mentions. Similarly, our c2f-coref system gets \textgreater96\% anaphoric mention recall yet only \textless40\% precision (Table~\ref{tab:original}). We therefore explore if detector recall is always more important than its precision. If more spans are considered by increasing the max span width $L$ or the number of spans considered per word $\lambda$, will the system performance necessarily improve? In the extreme case, if we hypothetically had enough compute that allows the linker to consider all $O(T^4)$ span-antecedent pairs, should we simply remove the pruning in the detector?

\paragraph{The Aggregated Importance of Precision}
For all oracles in Table~\ref{tab:oracles}, fixing precision yields a larger improvement than recall, especially with anaphoric mentions. %
This highlights the importance of detector precision and the extent to which the linker suffers from noisy mention candidates. In Table~\ref{tab:fix-errors}, we present the F$_1$ improvement after independently fixing categorized errors following~\citet{kummerfeld-klein-2013-error}.\footnote{Span (boundary) errors are fixed before independently fixing all others. The numbers do not add up to the performance gap due to error type interactions.} Noisy candidates result in extra mention and extra entity errors, fixing which accounts for more than half of the $\approx$8 F$_1$ gap between the post-hoc perfect precision oracle and the baseline for CoNLL-12 (Table~\ref{tab:oracles}). Furthermore, re-training the system to leverage the distributional shift of the absence of noise leads to another $\approx$4 and 5 F$_1$ increase (CoNLL-12/PreCo).

To analyze how higher detector precision helps the linker, we examine the coreference score the linker assigns to every span-antecedent pair.
The anaphoric mention re-trained perfect precision oracle has an average score of --13.0 on CoNLL-12, higher than --15.1 with perfect recall. Among only correct span-antecedent pairs, these scores are 11.7 and 7.1, with the same pattern. This indicates that the noise with perfect recall prevents the linker from reliably assigning high coreference scores, even for correct links. The effect of higher coreference scores also shows in that, compared with perfect recall, the perfect precision oracle produces on average larger (4.44 vs. 4.26 entities) and longer-distance (154 vs. 152 tokens spanned) clusters.

We also see this effect by examining the amount of improvement with reduced noise in Table~\ref{tab:fix-errors}. In the anaphoric mention post-hoc oracles, as expected, fixing precision results in fewer extra mention/entity errors and more missing errors, while the perfect recall oracle behaves conversely. However, when re-trained, the perfect precision oracle has much fewer missing entities, even fewer than with perfect recall. This is surprising as the latter considers more candidates. The reason is likely that the linker learns to leverage the absence of noise and reliably assigns high coreference scores. Despite some incorrect links leading to more conflated entities, the many correct ones drastically reduce missing mention/entity errors. On the other hand, the noise in the perfect recall or the original system prevents consistent high scores, resulting in more missing mentions and entities. Hence, the improvement with perfect precision partly stems from the linker's increased confidence in assigning coreference scores when \emph{not} tasked with ignoring non-mentions (and singletons) in noisy candidates.

\begin{table}[t]
\centering
\begin{tabular}{@{\hspace{4pt}}c|c|c|c|c@{\hspace{2pt}}}
\hline
\multicolumn{1}{c|}{} & \multicolumn{2}{c|}{CoNLL-12} & \multicolumn{2}{c}{PreCo} \\ \cline{2-5}
\multicolumn{1}{c|}{} & \# Op & Op effect & \# Op & Op effect \\ \hline
\nsm; P & 135.9 & 0.086 & 79.2 & 0.132 \\
\nsm; R & 1.9 & \textbf{0.715} & 0.8 & \textbf{0.709} \\ \hline
\am; P & 34.1 & \textbf{0.035} & 30.6 & 0.122 \\
\am; R & 115.3 & 0.002 & 4.1 & \textbf{0.143} \\ \hline

\end{tabular}
\vspace{-0.05cm}
\caption{\label{tab:operation-effects} The number of addition/removal operations needed for the oracle candidates, and the oracle performance increase, in F$_1$, amortized over each operation. Boldface indicates the higher per-operation effect between perfect precision and recall in each category.}
\vspace{-0.1cm}
\end{table}

\begin{table}[t]
\centering
\fontsize{9.5}{12} \selectfont
\begin{tabular}{@{\hspace{4pt}}c|c|c|c|c|c|c@{\hspace{4pt}}}

\hline
$L$ & 30 & 32 & 34 & 36 & 38 & 40 \\ \hline
F$_1$ & \textbf{79.17} & 78.76 & 78.86 & 79.03 & 78.88 & 78.85 \\ \hline \hline
$\lambda$ & 0.4 & 0.45 & 0.5 & 0.55 & 0.6 & 0.65 \\ \hline
F$_1$ & \textbf{79.17} & 78.99 & 79.15 & 78.35 & 79.05 & 78.85 \\ \hline

\end{tabular}
\vspace{-0.05cm}
\caption{\label{tab:more-spans} CoNLL-12 development F$_1$ with increased max span width $L$ or the number of spans considered per word $\lambda$. The first column is the original setting. Boldface indicates the best performance.}
\vspace{-0.3cm}
\end{table}

\paragraph{The Average Importance of Recall}
The large improvement from fixing precision may be due to its larger original headroom than recall (Table~\ref{tab:original}). We compute the number of operations (span addition/removal) needed for each oracle and the average F$_1$ improvement per operation in Table~\ref{tab:operation-effects}. For anaphoric mentions, recall has 5-8$\times$ the average effect of precision.\footnote{CoNLL-12 with all mentions has a different pattern as we noisily generated singletons in a recall-oriented way.} If we control the number of operations by re-training an anaphoric mention (semi-)perfect precision oracle removing only as many top-scoring extra spans as the number of missing correct spans (rather than removing all extra spans), it gets 79.08 and 85.01 F$_1$ on CoNLL-12 and PreCo, lower than the perfect recall oracle with 79.65 and 85.22. It is therefore only due to the low-precision high-recall nature of the original detector that precision is more important in aggregate.

\paragraph{Precision-Recall Trade-Off}
We return to the original question: if we had more compute, is it always beneficial to consider more spans in the detector? From our results, while recall is important, an imprecise detector has substantial adverse effects by increasing the linker's learning burden. Indeed, Table~\ref{tab:more-spans} shows that increasing the max span width by up to 33\% or the spans considered per word by up to 38\% only degrades the performance. As the extra low-scoring spans are mostly noise, we slightly increase recall but more heavily decrease precision, causing more harm than benefit. Hence, besides saving computation, these hyperparameters also balance the precision-recall trade-off. Future work should hence put more emphasis on precision which is often overlooked in end-to-end systems.

\section{Difficulties Facing Each Component}
\label{sec:focus}

\subsection{The Detector's Difficulty With Anaphoricity Decisions}
\label{sec:detector-difficulty}

Despite its large aggregated improvement, i.e. $\approx$11.7 and 10.5 F$_1$ for CoNLL-12 and PreCo, perfect anaphoric mention precision requires perfectly distinguishing anaphoric from singleton mentions. These anaphoricity decisions in fact account for most of the improvement, $\approx$10.5 and 6.7 F$_1$ (Table~\ref{tab:oracles}, anaphoric v.s. all mentions perfect precision).\footnote{\citet{chen-etal-2018-preco} observed a similar pattern on an LSTM architecture that less directly receives global information which is important for anaphoricity decisions. We confirm that this still holds on transformers with a larger receptive field.} However, the detector, as a span classifier, does not explicitly model inter-span anaphoric relationships. To test this architecture's ability to distinguish anaphoric from singleton mentions, we build two span classifiers with the same structure as the detector, supervised with sigmoid loss, that recognize all mentions and anaphoric mentions in PreCo. The former achieves 79.89 classification F$_1$ while the latter only 54.32, showing the inability of a span classifier to make anaphoricity decisions.

To better understand this difficulty, we define a confusion index as singleton recall divided by anaphoric mention recall. It correlates with the classifier's inability to identify anaphoricity. Ideally, this value should be close to 0, recalling more anaphoric mentions and fewer singletons. A random classifier incapable of distinguishing between the two has an expected confusion index of 1.

\begin{table}[t]
\centering
\fontsize{9.5}{11} \selectfont
\begin{tabular}{@{\hspace{1pt}} c@{\hspace{1pt}}|@{\hspace{3pt}}p{6cm} @{\hspace{2pt}}}
\hline
Error Type & \multicolumn{1}{c}{\multirow{2}{*}{Example}} \\
(\#) \\ \hline
\multirow{4}{*}{\shortstack{Pronoun\\(109)}} & ... a cross-sea bridge connecting \textbf{Hong Kong, Zhuhai, and Macao} . \\ \cline{2-2}
& ... after \textbf{their} return, Macao, and Hong Kong, the two special administrative regions ... \\ \hline
\multirow{4}{*}{\shortstack{Exact\\Match\\(6)}} & The most important thing about \textbf{Disney} is that it is a global brand . \\ \cline{2-2}
& The subway to \textbf{Disney} has already been constructed . \\ \hline
\multirow{4}{*}{\shortstack{Head\\Match\\(11)}} & \textbf{Ten landmark buildings located on Hong Kong Island} reveal themselves ...  \\ \cline{2-2}
& ... \textbf{those private , er , buildings} , that is , the business community , ah , is willing to ... \\ \hline
Other & And \textbf{Dr. Andy Henry} notices something else \\ \cline{2-2}
Match (7) & \textbf{Dr. Mann} says they 've narrowed it down ... \\ \hline
\multirow{4}{*}{\shortstack{Semantic\\Proximity\\(12)}} & ... \textbf{Hong Kong cinema} has nurtured many internationally renowned directors ... \\ \cline{2-2}
& ... memorializing \textbf{Hong Kong 's  100 - year film history} . \\ \hline
\multirow{2}{*}{\shortstack{Others\\(5)}} & But [\textbf{Paul Kelly}] [\textbf{Steve Sodbury}] and Mel Anderson ... had no idea ... \\ \hline

\end{tabular}

\caption{\label{tab:conflated-examples} Examples of categorized conflated entity errors in the CoNLL-12 development set with a perfect detector. Following past studies \cite{kummerfeld-klein-2013-error,joshi-etal-2019-bert}, we consider all deictic terms as pronouns. Each example contains two incorrectly linked entities in bold. Square brackets are added to separate mentions.}
\vspace{-0.3cm}
\end{table}

The anaphoric mention classifier above has a confusion index of 0.81, showing its inability to make anaphoricity decisions even when explicitly trained with the signal. If we only consider text appearing as both singleton and anaphoric mentions in the same document, demanding contextual reasoning by disregarding obvious anaphoric mentions such as pronouns, the confusion index degrades to 0.997. Hence, the classifier is poor at leveraging self-attentive contextual cues to make anaphoricity decisions without explicit inter-span relational modeling. In \S\ref{sec:confusion-index-vs-span-width} we also show the degradation of the confusion index with shorter spans.

Given the importance of anaphoric mention precision (\S\ref{sec:pr-re-analysis}), more research in improving anaphoricity decisions in the detector would be fruitful, for example, by more explicitly attending to neighboring spans. Alternatively, as \citet{zhong-chen-2021-frustratingly} showed the benefit of disentangling the span representations for entity detection and relation extraction in information extraction based on the intuition that they are disparate tasks, one may split the task of anaphoricity decision from mention linking and introduce a separately parameterized anaphoricity module, similarly considering the discrepancy between the two tasks. \citet{recasens-etal-2013-life,moosavi-strube-2016-search}; \textit{inter alia} have pursued similar ideas in the pre-neural era, but it has still not yet been explored with deep models.

\subsection{The Linker's Errors}

While the detector struggles with anaphoricity decisions, the linker explicitly models anaphoricity by assigning the dummy to extra mentions. %
It is hence also viable to determine anaphoricity in the linker.
Indeed, the current detector would suffice with a stronger linker: in Table~\ref{tab:oracles}, the oracle linker gets near-perfect scores with the original mentions (not perfect since the candidates are not gold).\footnote{A modified oracle linker that only considers coarse-pruned antecedents~\cite{lee-etal-2018-higher} still gets 96.61 and 98.65 F$_1$ on CoNLL-12 and PreCo. The small difference compared to considering all antecedents also shows that, with a strong linker, coarse-to-fine pruning has only negligible performance impact while substantially reducing the decision space.}

To analyze the remaining non-anaphoricity linker errors, we assume a perfect anaphoric mention detector. Here, conflated entities is the single major error source (last row of Table~\ref{tab:fix-errors}). Table~\ref{tab:conflated-examples} shows 150 manually categorized conflated entities in the CoNLL-12 development set.\footnote{\citet{joshi-etal-2019-bert} and \citet{lu-ng-2020-conundrums} conducted similar analyses but we study in a more controlled setting by excluding detector errors and focusing on entity conflation, the largest remaining error source.} Suboptimal pronoun resolution is the biggest issue, and the linker also tends to link spans with various degrees of text match or semantic proximity.
Within pronoun errors, the most common case is a pronoun linked to an incorrect nominal (in Table~\ref{tab:conflated-examples}), occurring 43 times. Sometimes two pronouns, often identical, are incorrectly linked, a case that necessitates better higher-order inference. Third person pronouns with different referents are conflated 29 times. Errors with first or second person pronouns occur 37 times, usually due to speaker switching.

Similar to \S\ref{sec:detector-difficulty}, separately parameterizing the linker's encoder may help reduce conflation: intuitively, the span representation for mention detection may promote homogeneity. Meanwhile, the lack of discerning span-internal content for certain error types including pronoun resolution and exact match, combined with current systems' trend to rely on such cues~\cite{lu-ng-2020-conundrums}, calls for more focus on improving their contextual reasoning.

\section{Conclusion}

We analyzed the complex interaction between the mention detector and linker in the mainstream coarse-to-fine coreference system. Using oracle experiments, we showed that, while detector recall is important, higher anaphoric mention precision would lead to dramatically better linker performance, though achieving this is difficult. We also demonstrated that the oracle linker performance is near perfect and that the vast majority of remaining linker errors besides anaphoricity decisions are about pronoun resolution. We hope these discoveries will help future coreference research.%

\bibliography{paper,anthology}
\bibliographystyle{acl_natbib}

\clearpage

\appendix

\section{Dataset Statistics}
\label{sec:dataset-statistics}

We use the English portion of the CoNLL-12 shared task~\cite{pradhan2012conll} and the PreCo~\cite{chen-etal-2018-preco} dataset. The former contains 2,802/343/348 training/development/testing documents and the latter has 36.6K training documents and 500 each for development and testing.

\section{Heuristically Generated CoNLL-12 All Mentions}
\label{sec:heuristics}

We heuristically generate the set of all mentions for CoNLL-12 in a recall-oriented manner. We use the gold syntactic information as a proxy and consider the union of all phrases tagged with NP or NML and all words tagged with PRP, PRP\$, WP, WDT, WRB, NNP, VB, VBD, VBN, VBG, VBZ, or VBP. This set includes 99.63\% anaphoric mentions which constitute 20.89\% of this set. We obtain the set of all mentions by merging this set with the non-singleton mentions to ensure all mentions are a superset of anaphoric mentions.

\begin{figure}[t]
    \centering
    \includegraphics[width=0.46\textwidth]{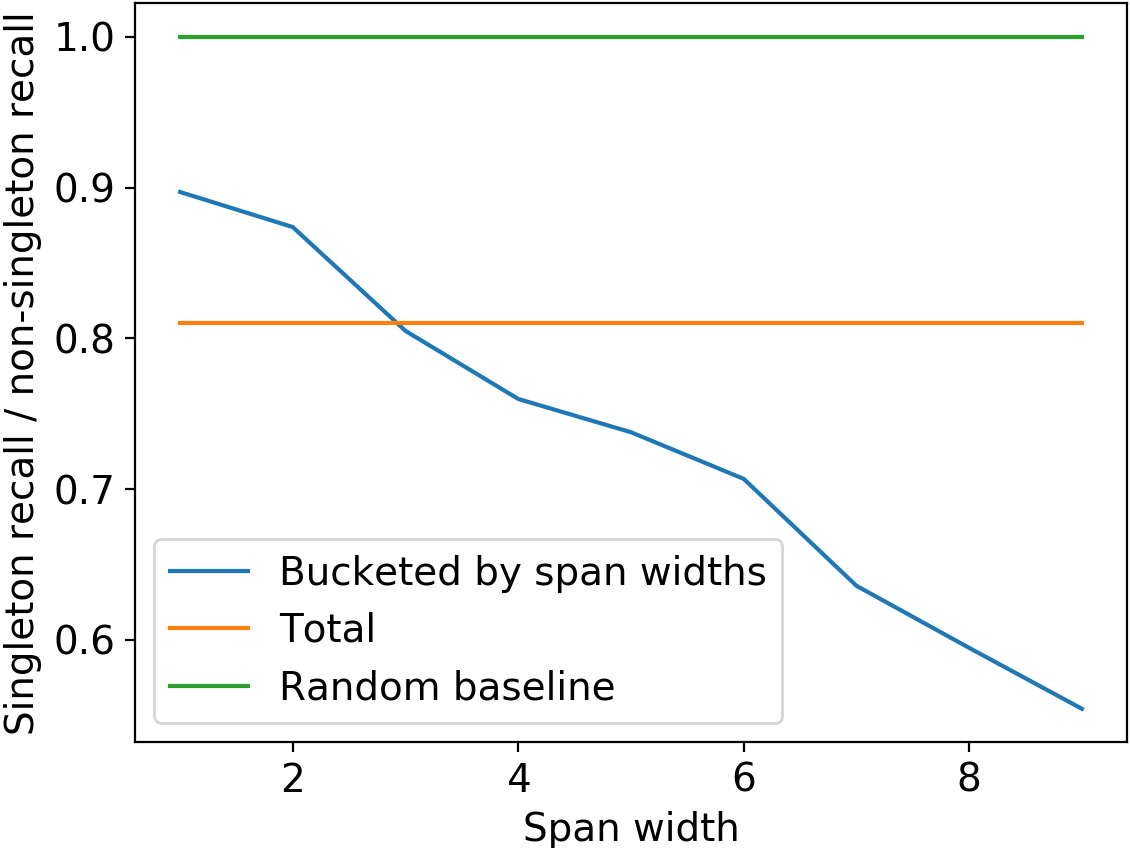}
    \caption{The PreCo anaphoric mention classifier confusion index (\S\ref{sec:detector-difficulty}) on spans with different widths.}
    \label{fig:confusion-vs-width}
\end{figure}

\section{The Confusion Index's Variation With Span Width}
\label{sec:confusion-index-vs-span-width}

In Figure~\ref{fig:confusion-vs-width}, we plot how the confusion index of the PreCo anaphoric mention classifier (\S\ref{sec:detector-difficulty}) changes with span widths. The classifier's inability to make anaphoricity decisions is the most pronounced for short phrases, possibly because these phrases are also more likely to appear as both singleton and anaphoric mentions whose anaphoricity status is especially hard to determine, discussed in~\S\ref{sec:detector-difficulty}.

\end{document}